# Prediction of Solar Radiation Based on Spatial and Temporal Embeddings for Solar Generation Forecast

Mohammad Alqudah[1], Tatjana Dokic[2], Mladen Kezunovic[2], Zoran Obradovic[1]

[1] Computer and Information Sciences Department
Temple University
Philadelphia, PA, U.S.A.

[2] Department of Electrical and Computer Engineering
Texas A&M University
College Station, TX, U.S.A

## Abstract

*A novel method for real-time solar generation forecast using weather data, while exploiting both spatial and temporal structural dependencies is proposed. The network observed over time is projected to a lower-dimensional representation where a variety of weather measurements are used to train a structured regression model while weather forecast is used at the inference stage. Experiments were conducted at 288 locations in the San Antonio, TX area on obtained from the National Solar Radiation Database. The model predicts solar irradiance with a good accuracy ($R^2$ 0.91 for the summer, 0.85 for the winter, and 0.89 for the global model). The best accuracy was obtained by the Random Forest Regressor. Multiple experiments were conducted to characterize influence of missing data and different time horizons providing evidence that the new algorithm is robust for data missing not only completely at random but also when the mechanism is spatial, and temporal.*

## 1. Introduction

Due to technological advances of solar power lowering the price of the photovoltaic (PV) panels and the push for cleaner energy, solar power has seen a tremendous growth worldwide. During the last decade the installed capacity for the number of OECD countries, all around the world has grown from 34% to 82% [1]. In 2017, renewables accounted for 55% of the 21 GW of U.S. capacity additions. Solar technology showed record 40% growth in power generation in 2017 [4]. As of February 2018, renewables accounted for 22% of total currently operating U.S. electricity generating capacity [2]. The tremendous growth in the U.S. solar industry is helping to pave the way to a cleaner, more sustainable energy future [3]. Furthermore, more solar plants are projected to be added to the power generation mix in the next few years.

With the rapid growth of the solar industry, the variability and intermittency of this renewable source of energy brings about major challenges in energy balancing which may affect the system reliability and flexibility. Since it can have a direct impact on consumers and businesses, it is very important to have an accurate real-time forecast of the solar generation so that both higher system operation efficiency and maximum solar utilization can be achieved [5].

Solar generation prediction techniques have been a research interest in the past few years. Type-1 and interval type-2 Takagi-Sugeno-Kang (TSK) fuzzy systems were proposed for the prediction of generation of solar power plants [6]. A multi-step scheme is developed to predict solar irradiance using weather data. A hybrid of the Autoregressive and Moving Average (ARMA) and the Time Delay Neural Network (TDNN) is applied in [7]. Numerical values of the atmospheric transparency index and the surface albedo from the NASA SSE database were used to develop the model for estimation of amount of solar radiation arriving at the arbitrarily oriented surface [8]. A promising model based on a vector autoregressive (VAR) framework fitted with two alternative methods (Recursive Least Squares and Gradient Boosting) is introduced [9]. An approach that uses classification, training, and forecasting stages is also proposed for 1-day ahead hourly forecasting of PV power output in [10]. First, the classification stage provides a self-organizing map (SOM) and learning vector quantization (LVQ) networks that classify the collected historical data. Then, the training stage employs the support vector regression (SVR) to train the input/output data sets for temperature, probability of precipitation, and solar irradiance of defined similar hours. Finally, in the forecasting stage, the fuzzy inference method is used to select an adequately trained model for accurate forecast. The multilayer perceptron (MLP), random forests (RF), k-nearest neighbors (kNN), and linear regression (LR), algorithms were used for solar irradiance forecasting [11].

Researchers started exploiting recently spatial correlations among geographically spread solar PV power plants, which led to improvements in the prediction accuracy [12-14]. In our previous study Gaussian Conditional Random Fields (GCRF) was used to forecast the solar power in electricity grids [5]. The introduced forecasting technique is capable of modeling both the spatial and temporal correlations of various solar generation stations.

Our paper introduces a novel prediction algorithm that combines the spatial and temporal embeddings and makes accurate predictions on multiple temporal horizons. The proposed method demonstrates good prediction accuracy, where $R^2$ of 0.91 is obtained for the summer model, 0.85 for the winter model, and 0.89 for the global model. Out of all the types of models that were tested (Linear Regression, Normalized Linear Regression, Support Vector Regression, Random Forest Regression, and Neural Networks), the best accuracy was achieved by Random Forest model. The robustness of the proposed algorithm was tested for different types of missing data cases (completely at random, spatial, and temporal) and the high accuracy is obtained in all of these instances.

The rest of the paper is organized as follows. Section 2 describes the background about solar generation forecast. Section 3 focuses on the prediction methodology. The results are presented in Section 4. The discussion and future work recommendations are in Section 5. Finally, Section 6 concludes the paper.

## 2. Background

Solar irradiance $I_{solar}$ is the power per unit area received from the Sun in the form of electromagnetic radiation in the wavelength range of the measuring instrument [14]. In solar power systems, the relationship between the solar power generation $P_{solar}$ and the solar irradiance $I_{solar}$ for a given material can be assumed as a linear relationship:

$$P_{solar} = I_{solar} \times S \times \eta \qquad (1)$$

where $I_{solar}$ is in ($kWh/m^2$); $S$ is the area of the solar panel in $m^2$; and $\eta$ is the generation efficiency of the solar panel material.

The high proliferation of PV generation in an electricity grid is challenging due to two main factors: variability and uncertainty [1]. Since it is highly dependent on weather conditions that are variable in nature, it can be hard to predict. This introduces a new challenge to the electric industry [15] compared with conventional power plants that are deterministically adjusted to the expected load profiles.

The amount of solar irradiance arriving at the solar panel depends on a variety of factors [8]. Some of the factors are deterministic and can be calculated using geometry, such as geographical location (latitude and longitude), and orientation angles of the solar panel relative to the Sun (declination angle, the hour angle, the zenith angle, the elevation angle, and the azimuth angle). Other types of factors are stochastic in nature. These include factors affecting the air between the solar panel and the Sun, such as concentration of atmospheric gases, dust, aerosols and water vapor suspended in the air, humidity, the nature of cloud cover, etc. While deterministic factors can be calculated for any location and any moment in time, stochastic parameters are obtained from the weather forecast for the future date and time.

Solar Zenith Angle (SZA) represents the angle between the Zenith and the center of the Sun's disc, where the Zenith represents an imaginary point directly over a particular location [16]. It has a high correlation with Global Horizontal Irradiation (GHI). The SZA is an important predictor of GHI and during the sunny days (without any clouds), SZA alone can be used to accurately estimate the solar irradiance. The SZA is a mathematically calculated value, it will be useful in any prediction model since it can be obtained without special equipment.

During the cloudy days, and especially during the days with high variability between sunny and cloudy intervals, the SZA is no longer enough for the accurate prediction of solar irradiance. In this case the stochastic parameters mentioned before have a major impact. Since these parameters are not deterministic, it is more challenging to provide accurate solar radiation forecast in the case of cloudy days.

In this paper we develop a data based prediction model for the forecast of the output power of the PV system using GHI, for a set of aggregated areas of a size 3 x 3 km. We use the National Solar radiation Database [17], and National Digital Forecast Database [18] to train and test the model.

We looked at different temporal horizons of PV forecast used in the industry [1]:

- *The day ahead (DA) forecast*. In this case the model based results are submitted the day before the operating day. The prediction is made for 24 hours, typically starting at midnight. Different utilities reported different times when the forecast is made, some make a forecast at 7 am the previous day and submit it at 9 am, others may submit the forecast at the end of a day shift at 5:30 pm. The last time point of forecast in this case could be larger than 24 hours in advance, sometimes up to 42 hours in advance.

The value of improvements in day-ahead forecasting is outlined in [19].

- *The hour ahead (HA) forecast.* This type of forecast is submitted 105 minutes prior to each operating hour. In some utilities it also contains an additional forecast for the next 8 hours of operation. We can conclude that this type of forecast predicts for a time horizon of 1.75 to 8.75 hours ahead. This type of forecast is a primary target of this paper.
- *Sub hourly forecast.* Utilities are also in the process of integrating the intra-hour forecast, going down to 5 minutes ahead. While our model is capable of addressing such forecast horizon as well, we are not focusing on this problem at this time.

PV forecast methods have different accuracy depending on the time horizon of forecast. Some methods perform better in a short term and some are better for a day-ahead forecast. Fig. 1 shows comparison of performance of different methods for a range of time horizons [20]. We can observe from Fig. 1 that cloud motion forecasts based on satellite (yellow and white lines) perform better than numerical weather prediction based on National Digital Forecast Database (NDFD) up to 5 hours ahead. Numerical weather prediction demonstrates similar forecast accuracy for time horizons going from 1 hour to 3 days ahead [1].

## 3. Methodology

In this section we describe the proposed data model used in the study. This model leverages the correlation between the locations where the data is collected with temporal weather data. This section first discusses the dataset used, and then introduces the proposed model.

### 3.1 Data

This research focuses on the problem how to leverage the correlations between spatial and temporal weather data to predict solar irradiation. As a result, the collected data has two parts: a network that represents spatial locations for the collected weather data, and temporal weather data.

**3.1.1 Spatial Data.** A set of locations ($L$) are used in this paper. Fig. 2 shows the 288 locations in the San Antonio, TX area where the data is collected. Each location $l_i$ in $L$ represents a 3 × 3 km area where solar irradiation is determined. For each location, the longitude and latitude are known which allows us to measure the distances between all the locations and build a spatial network. The built spatial network will be combined with the collected temporal data to make predictions for solar irradiation. This model is extracting the information that represents how different locations are affecting each other.

**3.1.2 Temporal Weather Data.** For each of the 288 locations discussed earlier, weather measurements are collected for the year (2017). In this data collection, weather measurements are collected every 30 minutes. In addition, solar irradiance collected by the National Solar Radiation Data Base [17] is spatiotemporally correlated with the weather measurements. The solar irradiance data also represents a measurement every 30 minutes for the same locations in San Antonio, TX. For each timestamp and in each location, the following weather measurements are collected: Dew Point, Solar Zenith Angle, Wind Speed, Precipitable Water, Wind Direction, Relative Humidity and Temperature. Since there is data each 30 minutes for 288 locations, the total number of data records is around 5 million.

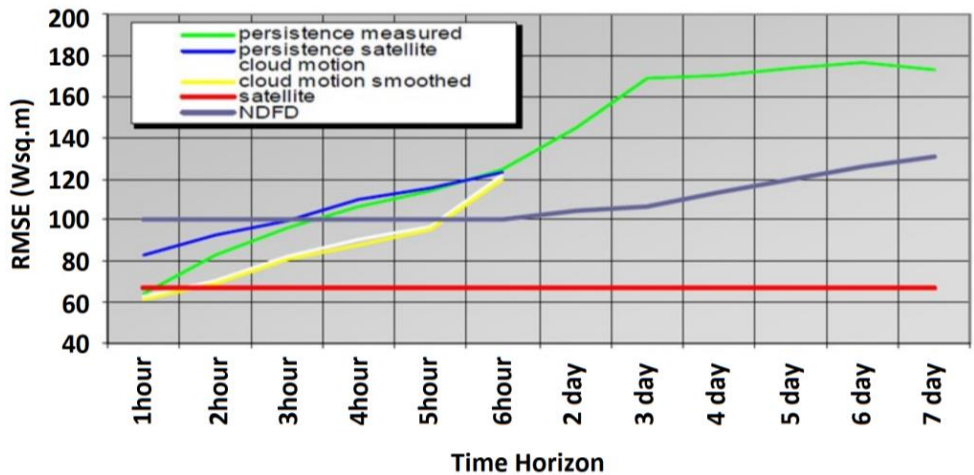

**Figure 1. Root mean square error (RMSE) of different solar forecasting techniques obtained over a year at seven SURFRAD ground measurement sites [20]. The red line shows the satellite nowcast for reference, i.e. the satellite 'forecast' for the time when the satellite image was taken.**

**3.1.3 Target Variable: Global Horizontal Irradiance (GHI).** Solar irradiance is represented by the Global Horizontal Irradiance (GHI) variable. This variable is collected every 30 minutes for all locations. For each of the 7 weather variables mentioned in Section 3.1.2, there is a GHI value corresponding to it.

**3.1.4 Correlation between Weather Variables and GHI.** There is significant correlation between some of the weather parameters and GHI. Table 1 shows the values of correlation. We can see that some of the parameters have high correlation, such as Solar Zenith Angle, and we expect those to be highly influential in the regression model.

### 3.2 Proposed Model

There are several methods of leveraging temporal and spatial data. In this model, we combine the temporal and spatial data by embedding the spatial information using Node2Vec [21], where the spatial correlations information is embedded to a new feature space ($S$). The temporal correlations are embedded by creating new features that represent the temporal correlations.

**3.2.1 Spatial Embedding.** This study considers spatial and temporal dependencies among 288 locations. Spatial dependencies of a certain site on remaining sites at a specific time can be represented as 574 variables corresponding to longitude and latitude for 287 locations. Data dimensionality is much larger when also considering temporal dependencies. Data sparsity in such a high high-dimensional representation is a major challenge for predictive modeling.

Another serious challenge is effective integration of relevant long-range spatial dependencies with local spatial information. A fusion of all available information can result in large data volume and large noise causing course of dimensionality and I/O problems, while too aggressive summarization can result in loss of important dependencies. One approach to address this challenge is to summarize the graph by aggregating locations of interest into "supernodes" representing larger regions. This can help reducing data dimensionality, but requires feature engineering which could cause additional challenges. Alternative methods, such as geographically weighted regression, were proposed to capture spatial interactions, but this is a serious challenge since a large spatial lag is problematic as it accounts for many irrelevant locations, while a small spatial lag ignores longer-range influences.

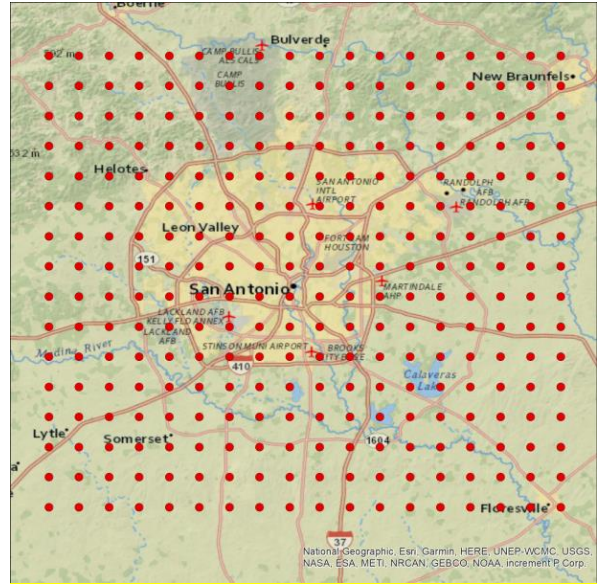

**Figure 2. Map showing the 288 locations where the temporal weather data is collected.**

**Table 1. Correlation coefficient between each weather parameter and GHI**

| Number | Feature Name | Correlation with GHI |
|---|---|---|
| 1 | Dew Point | -0.039 |
| 2 | Solar Zenith Angle | **-0.816** |
| 3 | Wind Speed | 0.296 |
| 4 | Precipitable Water | 0.017 |
| 5 | Wind Direction | -0.107 |
| 6 | Relative Humidity | -0.734 |
| 7 | Pressure | -0.105 |

Regression methods based on a naturally embedded spatial information (locations) typically assume spatial stationarity. For example, Autoregressive Statistical Methods adopt a spatial lag term to consider the autocorrelation of a neighborhood while geostatistical methods use semi-variograms to characterize the spatial heterogeneity. This assumption is another limitation since in practice relationships between variables vary at different locations.

An alternative and more flexible approach is to represent spatial data as a graph. This approach is appealing, but graphs can also add complexity to any learning model. Recently, progress was made in the representation learning field by embedding nodes of a graph or even entire graphs in a lower-dimensional space where standard machine learning methods could be used easier. The embeddings are learned and extracted by various algorithms. Such algorithms aim at conserving the graph structure and simplifying the learning models by moving away from graph representations. An advantage of using such methodologies is that they can potentially uncover more complex spatial dependencies that include some long-range interactions in addition to influences of the local neighborhood. An additional very useful property of using such approaches is that they can use jointly Euclidean spatial information and non-Euclidian variables.

The node embedding process represents the original graph in a new feature space $S$, where $S$ best-describes the spatial relationships of the nodes in the original graph. For instance, if two nodes $a$, $b$ depend on each other in the original graph (i.e.: geographically close, or longer-range dependent), this relationship will be retained when $a$, $b$ are represented in $S$. Hence $a$ and $b$ will be close in the new space $S$. This characteristic of the node embedding aims to capture essential relationships of the original graph structure while simplifying representation to a lower-dimensional list of feature vectors.

There are several algorithms to obtain such an embedding, two commonly used algorithms are DeepWalk [24] and Node2Vec [21]. Both algorithms rely on local information obtained by random walks used to learn latent space representations. A random walk (which is rooted at a vertex $v_k$) is a stochastic process with random variables $W_{v_k}^1, W_{v_k}^2, \ldots, W_{v_k}^j$, where $W_{v_k}^{j+1}$ is a vertex chosen randomly from the set of neighbors of vertex $v_k$. DeepWalk uses a stream of short random walks as the basis for extracting information from a graph by treating walks as the equivalent of sentences. DeepWalk is a generalization of a language model aimed to explore the graph through walks. In this analogy, the walks can be considered phrases in a special language.

In addition to the ability of DeepWalk to capture community information, DeepWalk is able to perform local exploration efficiently and is able to accommodate small changes in the graph structure without global recomputation. DeepWalk has two main steps: random walk generator and update procedure. In the random walk, a vertex $v_k$ is uniformly sampled from the graph and set as the root of the random walk. Then a walk uniformly samples from the neighbors of the last visited vertex until the maximum length of the walk is reached. In the second step, the update procedure called SkipGram updates the representations in accordance to the defined objective function. SkipGram is a language model that maximizes the co-occurrence probability among the word within a window in a sentence. Since the random walks of a graph can be considered phrases in a sentence, SkipGram will maximize the probability of its neighbors in the walk. The final representation is obtained through a hierarchical softmax process. DeepWalk includes optimization and parallelizability features, which allows to obtain a good performing representation (against a target function) though 32-64 random walks of a window width of 40.

Node2Vec is an algorithmic framework that generalizes DeepWalk process to provide a flexible notion of a node's neighborhood which allows learning richer representations by effectively exploring diverse neighborhoods. Node2Vec achieves better representations by introducing a search bias $\alpha$ to its random walks. This allows Node2Vec to explore different types of network neighborhoods. $\alpha$ allows discovering short and long distances by incorporating two parameters $p, q$ which guide the walk. The return parameter $p$ controls the likelihood of immediately revisiting a node in the walk while $q$, the in-out parameter, allows the search to differentiate between inward and outward nodes. Unlike DeepWalk, Node2Vec is sensitive to neighborhood connectivity patterns in networks.

In Node2Vec, as well as DeepWalk, the number of output dimensions is a hyperparameter and must be predefined. Using low setting for the number of dimensions (<16) can affect the stability of the generated space [24]. The literature shows the values for the dimensions to be the most effective if set between 32 and 128, in integer increments to the power of 2.

To convert the dataset in Fig. 2 using Node2Vec, a connected graph is created from the 288 locations. In order to achieve this, distances between locations are used as edge weights in a fully connected graph. Then, Node2Vec is used to convert Fig. 2 to the new feature space $S$, where each location $l_i$ in $L$ is mapped to a vector $s_i$ in the embedding space $S$. The final conversion is a matrix of size $288 \times D$ where $D$ is the number of dimensions for S. In Node2Vec, $D$ is a hyper-parameter that needs to be determined in advance.

**3.2.2 Temporal Embedding.** In order to preserve the temporal relationship included in the collected data, temporal embeddings ($T_E$) are used. Temporal embeddings can be useful in modeling temporal data, but one must be careful not to over-embed the data. This might cause the model to rely on the temporal aspect to learn the target variable, and this might lead to over-fitting. In this model, two embeddings are created for the time period $T$ and used in the model: Hour of the day and Season. For hour of the day, it is a simple 0 to 23 value of the hours in the day. For the season (winter, spring, summer and fall), it is determined from the date of each measurement taking in consideration the leap years.

Fig. 3 shows the shape of the dataset after the embeddings. Spatial embeddings S are added in addition to the temporal embeddings and weather attributes. The final embedded data $E$ is a concatenation of , weather attributes and $T_E$.

**3.2.3 Data Aggregation and Model Flow.** Fig. 3 represents how the data is aggregated and used in training the regressor. The ID column represents the location ID. Time is a sequential counter. ID and Time are not used in any learning and are shown here for demonstration purposes only. One can see that for each location and each reading, spatial and temporal embeddings are appended. In the original dataset, a very small number of records (< 0:001%) had missing values, and those records were dropped from the dataset. The dataset is temporally split for training and testing. Section 4.1 discusses how the data is split for training and testing purposes.

**3.2.4 Regression Model.** After data is transformed to the shape showed in Fig. 3, a regression model that predicts a value for GHI based on the input ($E$) is built. In this study, several regression models were tested, including Linear Regression, Normalized Linear Regression (Ridge, Lasso), Support Vector Regression (rBf kernel, linear kernel), Random Forest Regression, and Neural Networks.

The best accuracy was obtained by the Random Forest (RF) Regression [22]. RF is an ensemble of tree predictors, where each tree depends on an independent and randomly sampled vector with the same distribution as all the other trees in the forest. RF is an ensemble of B trees $\{T_1(X), \ldots, T_B(X)\}$, where $X = \{x_1, \ldots, x_p\}$ are independent and randomly sampled vectors with the same distribution. The ensemble of B trees produces B outputs $\{\widehat{Y_1} = T_1(X), \ldots, \widehat{Y_B} = T_B(X)\}$ where $\widehat{Y_b}$ is the prediction of the $b$th tree. The final aggregation of the regression is an average of the individual tree predictions.

## 4. Results

### 4.1 Temporal Data Split

The dataset described earlier has a strong temporal factor embedded in it. Thus, we utilized temporal hold-out validation is reserved for validation instead of relying on k-fold cross-validation. Following models were trained and tested:

1. *Winter model*: using October and November data for training and using December data for testing.
2. *Summer model*: using June and July data for training and using August data for testing.
3. *Global model*: using December and August data for testing and the remaining months for training.

The rationale behind this split is the following: for Summer and Winter models, we expect to have close correlation for these specific months, since the weather is somehow similar; for example, during the summer months we expect a large number of sunny days, while during the winter months we expect higher number of cloudy and variable days. For the global model, the idea is to test a generalized model using all data, and see if combination of the Winter and Summer models produces a good performing global model.

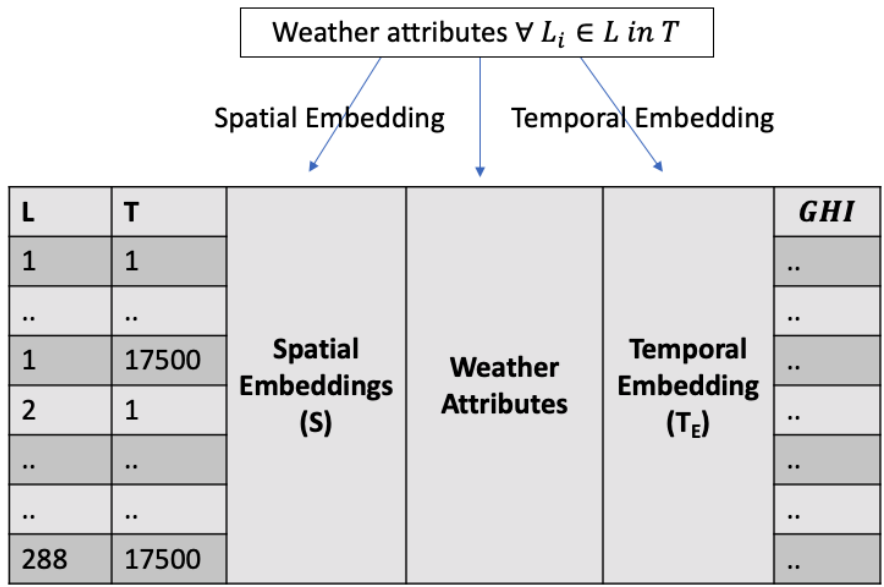

**Figure 3. Shape of the dataset after the embedding process**

### 4.2 Test Data

The three models described in section 4.1 use weather measurements for training, which means that all months of the year except August and December are *actual* weather measurements. In order to have a realistic test data, the testing months (August and December) are using *weather predictions* made 3 hours in advance before the timestamp of the GHI measurements. This experimental setup ensures that the model is testing a scenario that is very similar to a real-world application.

### 4.3 Data Preprocessing

Original weather data contains a very minimal set of missing values (0.01%). The records having missing values are removed since the dataset is large and the amount of missing data is insignificant. After removing missing values, the embedded dataset $E$ is constructed. Spatial embedding $S$ is constructed using $D = 32$. Furthermore, temporal embeddings $T_E$ are constructed. Since temporal embeddings introduces a categorical variable, 1-hot encoding is used to encode the categorical variables into a bit-vector where a single digit is 1 corresponding to a specific category. The last step is to scale $E$ to a $[0, 1]$ scale using a min-max scaler. This is necessary to remove the effects of different data scales for different variables. Both operations were conducted using Scikit-learn preprocessing module [23].

### 4.4 Model Training and Results

Model test results were evaluated by using coefficient of determination denoted as $R^2$, mean absolute error (MAE), and root mean squared error (RMSE). Best performing regressors was Random Forest Regressor, with $R^2 = 0.91$, MAE = 42.76, RMSE = 92.8.

#### 4.4.1 Summer Model.
In this model, weather measurements from June and July of 2017 are used for training while weather predictions for August 2017 are used for testing. Table 2 shows the results for the summer model. As expected, the summer model has good performance, since summer months usually have lower variation in the weather, hence more predictability of GHI. Fig. 4 shows the predicted GHI using the summer model for 100 readings.

#### 4.4.2 Winter Model.
In this model weather measurements from October and November of 2017 are used for training. The weather predictions for December 2017 are used for testing. The summer model performs better than the winter model. This is expected due to the higher number of clear sunny days in the summer when the correlation between SZA and GHI is very high as explained in Sec. 2. Table 3 shows the results for the winter model.

#### 4.4.3 Global Model.
In this model weather measurements from December and August of 2017 are used for testing and the rest of the months of 2017 are used for training. As expected, this model performs better than the winter model. Table 4 shows the results for the global model.

Table 2. Predictions 3 hours ahead by the summer model

| Metric | $R^2$ | MAE | RMSE |
|---|---|---|---|
| Value | 0.91 | 42.76 | 92.8 |

Table 3. Predictions 3 hours ahead by the winter model

| Metric | $R^2$ | MAE | RMSE |
|---|---|---|---|
| Value | 0.85 | 27.3 | 71.49 |

Table 4. Predictions 3 hours ahead by the global model

| Metric | $R^2$ | MAE | RMSE |
|---|---|---|---|
| Value | 0.89 | 33.4 | 85.2 |

### 4.5 Spatial Embedding Sensitivity Study

There is one hyper parameter $D$ which represents the number of dimensions in the spatial embedding. Typically, in embedding dimensions several values of the power of 2 are tested (32, 64, etc.). In this experiment 32, 64 and the default Node2Vec 128 are tested. There were no significant differences between the dimensions, and this can be interpreted as the graph being a symmetrical static grid. Also, another variation of the graph is embedded, which was created by dropping the top 10% of the links (distance-wise). This variation didn't make a difference in the results, and it was similar to the results reported earlier.

### 4.6 Handling Missing Data

Missing data is one of the common problems seen in this domain. In the following set of experiments, few scenarios of missing data are simulated and tested.

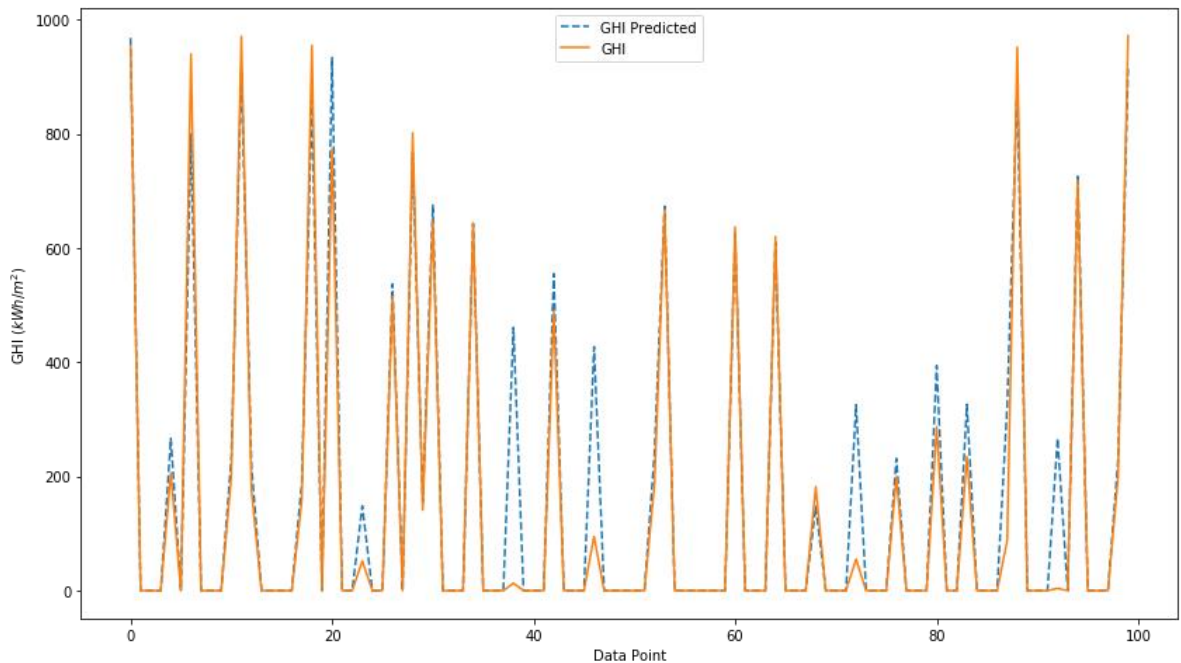

**Figure 4. GHI vs. GHI predicted 3 hours ahead for 100 readings using the summer model.**

1. Random Missing Data: in this experiment data is dropped from the dataset completely at randomly. An experiment to drop 30%, 50%, 70%, 90% and 95% of the training data. Given that the training set has 3.2 million records, it is expected that dropping data randomly will not be highly effective. All models performed almost the same. Even the model running on 99% performed similarly since 99% is around 33K rows of data, which seems to be enough to train this model. This will be discussed more in Section 5.
2. Spatial Missing Data: in this experiment, random locations are dropped completely from the training data. In this experiment, 10, 20, 30, 50, 75, 100 and 150 locations are dropped completely from the training dataset. We still used those locations in the testing dataset (testing data didn't change). Similar to point 1, the model performed similarly. This might be due to proximity of the locations. This will also be discussed more in Section 5.
3. Temporal Missing Data: This experiment is conducted in the following ways:
   - Removing a season from the training data (in the global model, example: dropping all data from the Spring season): this didn't affect the model. The model behaved similarly.
   - Lowering the resolution of the training data: the original training dataset has readings every 30 minutes. In this experiment, the resolution of the data is lowered to 1, 2, 4, and 8 hours. The model behaved similarly with some slight decline.

The results reported in this section provide evidence that the model robust to data missing at various mechanisms. Extending the cases of missing values is out of scope of this research. Again, more discussion is given in Section 5.

### 4.7 Weather Feature Importance

One advantage of using Decision Tree Regressor is that it produces (by default) feature importance for the features used, which could be used as feature ranking [23]. Fig. 5 shows feature importance for the top 15 parameters extracted from the Decision Tree Regressor trained in the embedded representation. As expected, Solar Zenith Angle is the most important feature, then humidity and perceptible water are the next important features. As we discussed earlier, this is expected as Solar Zenith Angle is directly related to the amount of solar radiation for the sunny days without clouds. In addition, humidity and perceptible water can directly affect how the Sun radiation affects an area, which has a strong relation to GHI.

### 4.8 Evaluating Longer Prediction Horizon

Results reported in previous sections were for a 3-hour horizon. Figure 6 shows prediction accuracy of the proposed method predicting 3, 6, and 9 hours ahead. The results are reported for winter, summer and global models. All 3 models show good stability over the longer horizons. Global model has mean values of ($R^2$=0.87, MAE=45.7, RMSE=100.6). In comparison, summer model has mean values of $R^2$=0.86, MAE=62.2, RMSE=122.1 and winter model has mean values of $R^2$=0.83, MAE=36.7, RMSE=87.1). This shows that the results are consistent across different time horizons. A slight improvement in RMSE was observed for the winter model at 9-hour horizon since predicting GHI is less complex near the end of the daytime.

## 5. Discussion and Future Experiments

Several aspects about the data generation process and results deserve a discussion. The dataset is constructed as a combination of satellite data, radar data, and mathematical models. This might explain the strong performance of the proposed model even when trained on a small data subset. On the other hand, GHI has temporal patterns (low in the morning, peak during the day, and then declines, and goes to 0 at night). This is one of the factors that can help in improving the performance of the model. Also, this study uses a large dataset, and this helped improve the results. Another factor is that in conducted experiments locations were not far from each other. From Fig. 2, the width of the grid is less than 35 miles, which makes the weather pattern similar in these locations. Temporal embeddings are helping the model and not over-fitting the data. An experiment conducted without 'time of day' embedding showed that the model is not learning the GHI by time only.

## 6. Conclusion

Following are the contributions of the paper:
- A novel approach to solar radiation forecast is developed based on spatial and temporal

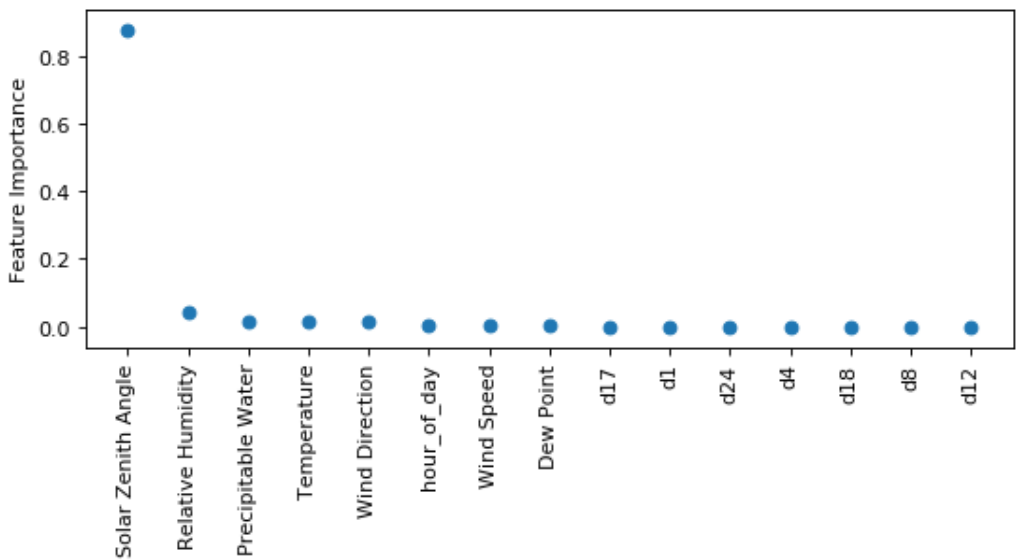

Figure 5. Feature importance

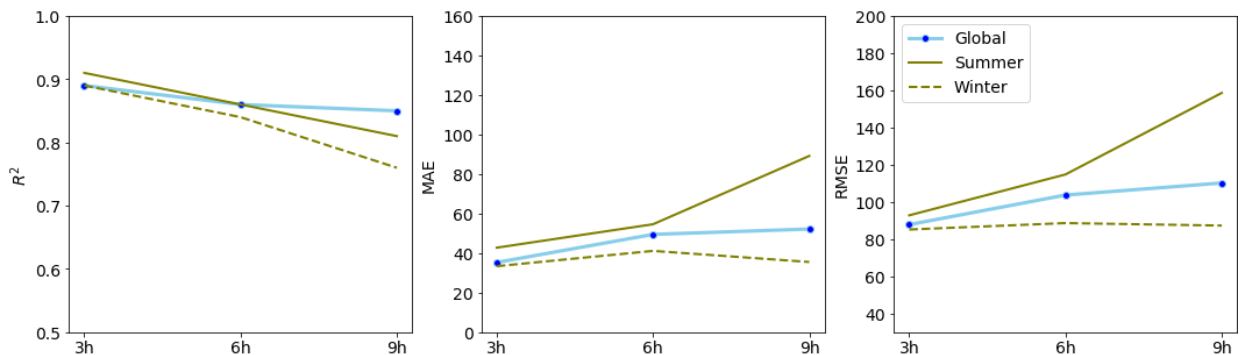

**Figure 6: Prediction accuracy 3, 6 and 9 hours ahead by global, summer and winter models**

embeddings using Node2Vec model for graph data. This approach simplifies the learning models by moving away from complex graphs. The model was developed for the forecast ranging from 3 to 12 hours ahead.
- The performance was tested for multiple regression algorithms: Linear Regression, Normalized Linear Regression, Support Vector Regression, Random Forest Regression, and Neural Networks. The Random Forest Regression has shown the best results.
- Variability of prediction accuracy for different seasons was explored. As expected, the algorithm performed with a very high accuracy in the summer when there are more clear sky days. During the winter months, the accuracy had a slight drop, but was still good and robust even when data was missing spatially and temporally.